\documentclass{article}

\PassOptionsToPackage{sort&compress,square,numbers}{natbib}

\usepackage[preprint]{neurips_2025}

\usepackage[utf8]{inputenc} 
\usepackage[T1]{fontenc}    
\usepackage{hyperref}       
\usepackage{url}            
\usepackage{booktabs}       
\usepackage{amsfonts}       
\usepackage{nicefrac}       
\usepackage{microtype}      
\usepackage{xcolor}         
\usepackage{graphicx}
\usepackage{amsmath}
\usepackage{cleveref}
\usepackage{colortbl,multirow,float,xcolor,algorithm,algpseudocode,amsmath}
\usepackage{changepage}
\usepackage{wrapfig}
\usepackage{caption}
\usepackage{enumitem}
\usepackage{bm}
\definecolor{positive}{HTML}{E5F6DA}
\definecolor{neutral}{HTML}{FBEADA}
\definecolor{negative}{HTML}{FADCDB}

\title{Real-Time Verification of Embodied Reasoning for Generative Skill Acquisition}

\author{%
  Bo Yue${}^{1\dagger}$, Shuqi Guo${}^{1\dagger}$, Kaiyu Hu${}^{1}$, Chujiao Wang${}^{1}$, \\
  \textbf{Benyou Wang${}^{1}$, Kui Jia${}^{1}$, Guiliang Liu${}^{1}$\thanks{Corresponding to liuguiliang@cuhk.edu.cn}} \\
  ${}^{1}$School of Data Science, The Chinese University of Hong Kong, Shenzhen\\
  ${}^{}\dagger$ Equal contribution
}

\begin{document}

\maketitle

\begin{abstract}
Generative skill acquisition enables embodied agents to actively learn a scalable and evolving repertoire of control skills, crucial for the advancement of large decision models. While prior approaches often rely on supervision signals from generalist agents (e.g., LLMs), their effectiveness in complex 3D environments remains unclear; exhaustive evaluation incurs substantial computational costs, significantly hindering the efficiency of skill learning. 
Inspired by recent successes in verification models for mathematical reasoning, we propose VERGSA 
(\underline{V}erifying \underline{E}mbodied \underline{R}easoning in \underline{G}enerative \underline{S}kill \underline{A}cquisition),
a framework that systematically integrates real-time verification principles into embodied skill learning. VERGSA establishes 1) a seamless extension from verification of mathematical reasoning into embodied learning by dynamically incorporating contextually relevant tasks into prompts and defining success metrics for both subtasks and overall tasks, and 2) an automated, scalable reward labeling scheme that synthesizes dense reward signals by iteratively finalizing the contribution of scene configuration and subtask learning to overall skill acquisition. To the best of our knowledge, this approach constitutes the first comprehensive training dataset for verification-driven generative skill acquisition, eliminating arduous manual reward engineering. Experiments validate the efficacy of our approach: 1) the exemplar task pool improves the average task success rates by $21\%$, 2) our verification model boosts success rates by $24\%$ for novel tasks and $36\%$ for encountered tasks, and 3) outperforms LLM-as-a-Judge baselines in verification quality.
\end{abstract}

\section{Introduction}\label{sec:intro}

Recent advances in Embodied AI have highlighted generative skill acquisition as a key research focus, driven by its ability to facilitate continuous learning, adaptability in dynamic environments, and scalable decision-making. By empowering large decision models to autonomously acquire and refine skills, Embodied AI systems can tackle an expanding range of diverse tasks, optimize decision-making processes, and ultimately improve their performance in complex, real-time scenarios. Reasoning plays a crucial role in this paradigm, providing structured supervisory signals that guide skill learning and ensure that acquired skills are aligned with task objectives.

As a pioneering enabler of advanced reasoning, Large Language Models (LLMs) have proven highly effective across diverse decision-making domains 
\cite{kaddour2023challenges,shao2024deepseekmath,ma2024survey,wu2024survey}. Despite their success, even state-of-the-art LLMs struggle with long-horizon, multi-step reasoning tasks \cite{huang2022towards}. Recent advances in verification models alleviated part of the issue, particularly in the context of mathematical problem-solving \cite{cobbe2021training,uesato2022solving,li2023making,lightman2023let,wang2024math,zhang2025lessons,setlur2025rewarding}, where a verification model offers a reliable benchmark for assessing candidate responses, thereby facilitating LLMs' self-improvement of reasoning capabilities.

\begin{wrapfigure}{t}{0.5\textwidth}
    \vspace{-12pt}
    \includegraphics[width=0.5\textwidth]{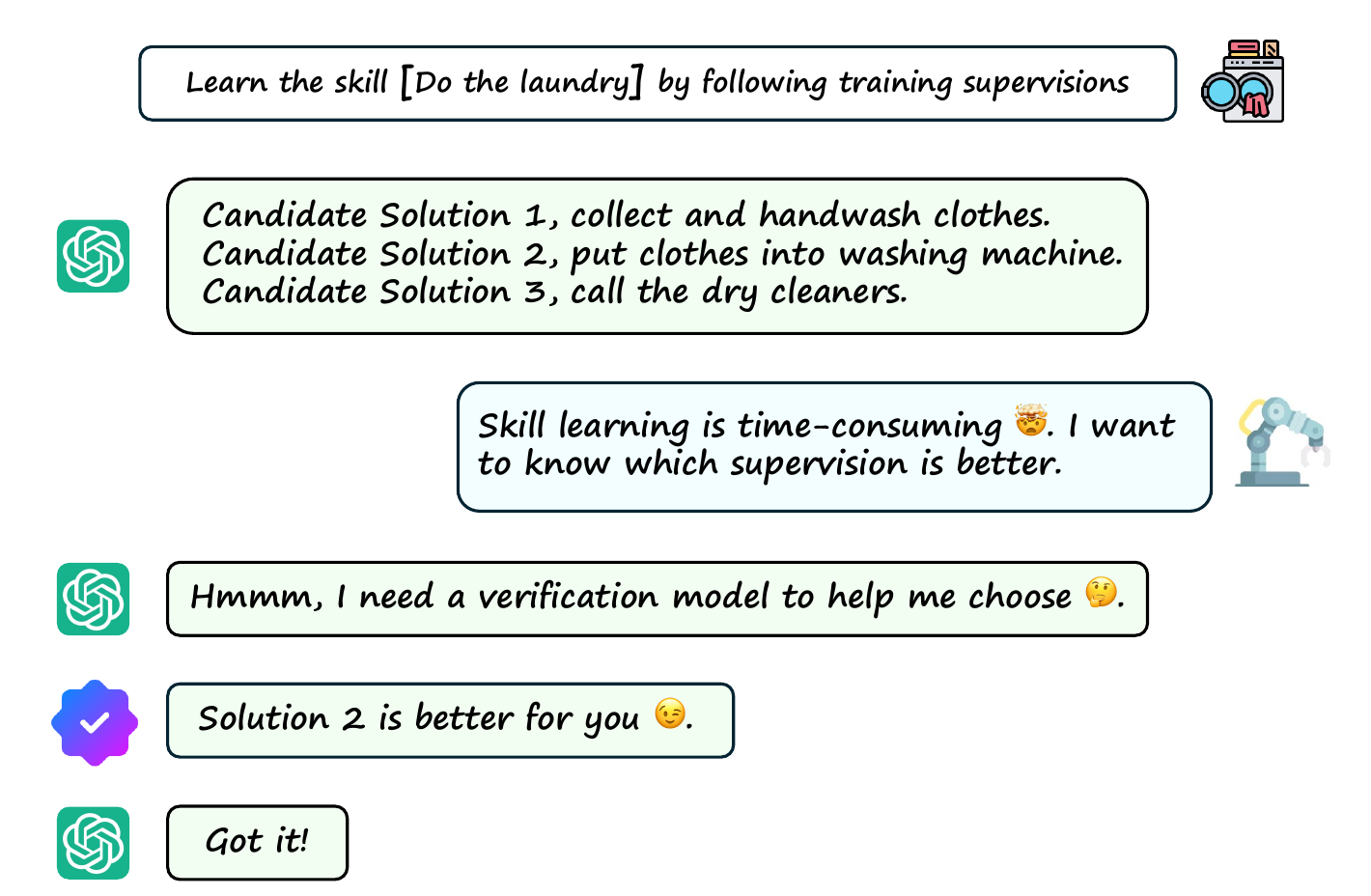}
    \caption{A motivating example.}
    \label{fig:motivative-example}
    \vspace{-8pt}
\end{wrapfigure}

Given the substantial time and computational resources required for embodied skill learning, exhaustively iterating every candidate solution is neither efficient nor practical. Furthermore, this approach offers limited guidance for tackling novel tasks. Inspired by the role of verification models in mathematical problem-solving, we investigate the potential utilization of such models in the context of embodied reasoning for generative skill acquisition. A motivating example is shown in Figure \ref{fig:motivative-example}, where the verification model reranks them to accelerate and enhance generative skill acquisition.

We observe that embodied tasks differ from mathematical problems in two critical aspects:
{\it 1) Less Structured}: While mathematical problems can be represented and solved using well-established symbolic formulas, robotic tasks are inherently less structured. They require configuring the scene, adaptively decomposing long-term goals into sequentially coherent subtasks, generating training supervisions and success metrics to guide the learning process of embodied agents.
{\it 2) Verification Challenges}: Unlike mathematical problems, which have definitive and directly verifiable answers, evaluating the success of skill learning in embodied tasks requires an external model to observe and assess performance.
In light of these distinctions, a crucial question arises: 
{\it `How can we optimize generative skill acquisition by integrating the real-time verification model into embodied tasks?'} 

To answer this question, we propose a framework for 
\underline{V}erifying \underline{E}mbodied \underline{R}easoning in \underline{G}enerative \underline{S}kill \underline{A}cquisition (VERGSA). 
Our approach comprises two key components: 1) a seamless extension for verification from mathematical problems into embodied tasks and 2) a scalable training procedure of process verifiers.

Firstly, we leverage contextually similar, successfully executed tasks as reference knowledge for novel embodied tasks to bridge the `less structured' gap. 
A dynamic exemplar task pool curates structured representations of both historical and incrementally added future tasks with successful outcomes. Additionally, to automate skill evaluation, we prompt the LLM to generate success metrics as part of the training supervision for each subtask to address the `verification challenges' gap.

Secondly, to train a verifier, VERGSA introduces a scalable labeling scheme that learns a Process Reward Model (PRM) to guide intermediate reasoning steps in embodied tasks.
Prior work relied on extensive human annotation to label rewards for steps in mathematical problems \cite{uesato2022solving,lightman2023let}, a challenge further amplified in embodied tasks due to the high cost of diverse, domain-specific expertise.
To address this issue,  VERGSA leverages Monte Carlo Tree Search (MCTS) as the reasoning policy to simulate consecutive subtask sequences with supervisions that completes the skill acquisition, starting from a specific substep. Rewards are assigned based on each substep's contribution to successful skill completion, enabling automatic estimation of reward signals from execution feedback.

Finally, we employ the well-trained verifier to guide the policy model, producing refined scene configurations and subtask supervisions to accommodate future skill acquisition tasks. Experiments have validated that: 1) the exemplar task pool offers important guidance that elevates average task success rate by $21\%$; 2) the verification model boosts success rates by $24\%$ for novel tasks and $36\%$ for encountered tasks; and 3) outperforms LLM-as-a-Judge with leading LLMs as verifiers.

We summarize our {\bf main contributions} as follows:
\begin{itemize}
    \item We propose a novel VERGSA framework that effectively grounds LLM reasoning capabilities in embodied tasks through a scalable and generalizable design;
    \item To the best of our knowledge, we construct the first embodied reasoning verification dataset with detailed scene configurations and subtask-level training supervisions, which applies to various downstream training scenarios;
    \item We develop an effective verification model that facilitates generative skill acquisition.
\end{itemize}

\section{Related Works}
\noindent{\bf Process Reward Model.} A verification model ensures consistency in LLMs' multi-step outputs, offering crucial feedback for improvement.
There are two main types of verification models: the Outcome Reward Model (ORM) and the Process Reward Model (PRM). The ORM assigns a reward to the entire solution, whereas the PRM allocates rewards to each intermediate substep based on its contribution to the final correct answer. As a result, PRMs are particularly well-suited for tasks that require complex reasoning, where substep reward signals reflect the nuances of the decision-making process.
PRMs have primarily been studied in the context of multi-step mathematical reasoning problems \cite{cobbe2021training}. Early research focused on training PRMs for mathematics using step-level human-annotated data \cite{uesato2022solving, lightman2023let}. More recent efforts have shifted towards developing automated pipelines for generating labels to train datasets \cite{wang2024math, zhang2025rest, setlur2025rewarding, li2025process}. However, the application of PRMs to autonomous skill acquisition remains largely unexplored. While these embodied tasks also require reasoning, they differ from mathematical problems in two key ways: 1) a scene configuration should be defined for a task; 2) each subtask should be equipped with training supervision to guide embodied agents to learn competent policies, and 3) context-dependent metrics should be designed to validate the success of an embodied task and subtasks within it.

\noindent{\bf Autonomous Skill Acquisition.}
Scalable robotic skill acquisition with minimal human supervision has become a prominent topic in Embodied AI. Prior work leveraged LLMs to reasonably stack skills from a basic library to perform complex, long-horizon tasks \cite{zhang2023bootstrap,du2023guiding}. Recent breakthroughs in the multi-modal foundation and generative models \cite{poole2022dreamfusion,melas2023realfusion,driess2023palm,achiam2023gpt,touvron2023llama} have further propelled this field, facilitating the development of essential building blocks for skill acquisition, such as goal specification \cite{kapelyukh2023dall,jiang2023vima}, scene construction \cite{xiang2020sapien,james2020rlbench,deitke2023objaverse,liu2023zero}, sub-task planning \cite{ahn2022can,huang2022inner,lin2023text2motion,ha2023scaling}, code generation \cite{wu2023tidybot,liang2023code,yu2023language,ma2023eureka}, data augmentation \cite{yu2023scaling}, and skill generalization \cite{brohan2023rt}. Among the ongoing efforts integrating these components for scalable skill learning \cite{wang2023gensim,katara2024gen2sim}, RoboGen \cite{wang2024robogen} stands out by introducing an automated pipeline that leverages the generative and commonsense capabilities of foundation models to generate task, scene, and training supervision. However, it does not verify either scene configurations or subtask supervisions, crucial for ensuring robust and efficient embodied skill learning.

\vspace{-0.12in}
\section{Problem Formulation}
\vspace{-0.12in}
\emph{Embodied reasoning for generative skill acquisition} consists of two primary components: 1) \emph{scene configuration}: generation of the interaction environment, including embodiments and scene objects, 
based on a given task (name and description); 2) \emph{skill generation}: generation of a sequence of decomposed subtasks, along with training supervisions and success metrics to learn low-level policy for each subtask, based on the generated scene. 
We formalize this process as a Markov Decision Process (MDP), denoted by $\mathcal{M}=(\mathcal{S},\mathcal{A}, P_\mathcal{T}, r, \gamma, \mu_0)$, where $\mathcal{S}$ and $\mathcal{A}$ are the state and action spaces; $P_\mathcal{T}$ and $r$ are dynamics and rewards; $\gamma\in(0,1]$ is the discount factor; The initial state $\mu_0$ is defined as the input task $g=(\text{name, description})$ that is drawn from a distribution of embodied tasks \({G}\), i.e., $g \sim G$.
A frozen policy model \(\pi\in\Delta_{\mathcal{S}}^{\mathcal{A}}\) samples an action $a \in \mathcal{A}$ based on a given state.
Each state is a concatenation of past actions, i.e., $s_t=\oplus^{t-1}_{i=1} a_i$.
The policy model may be any pre-trained LLM with reasoning capabilities. The $k$-th of $K$ complete reasoning trace $(c_k, z_{k,1},\dots,z_{k,T_k})\in\mathcal{A}^{T_k+1}, k\in[K]$ consists of the scene configuration $c_k\in\mathcal{A}$ and a sequence of subtask supervisions $(z_{k,1},\dots,z_{k,T_k})\in\mathcal{A}^{T_k}$. Here and in the following, we slightly abuse the term `subtask supervision' to include decomposed tasks, training supervision, and success metrics mentioned in the skill generation part. Unlike reasoning in mathematical problem-solving, where a problem (scene) is given, we reason through each intermediate step in the complete generative skill acquisition pipeline, including generation of scene configurations. Specifically, reasoning is achieved by a process verifier $V_{\theta}: {G}\times\Omega \mapsto [0,1]$, which estimates the expected task success probability for a \emph{real-time} generated trace prefix: $V_{\theta}(g, \Phi^g_k)\approx\Pr\bigl(r^g_k = 1 \mid g,\Phi^g_k\bigr)$. 
Each trace prefix is drawn from a set of possible trace prefixes $\Omega(g)=\Bigl\{
\Phi^g_k:=(c_k, z_{k,1},\dots,z_{k,t_k})\in\mathcal{A}^{t_k}
\,\Big|\,
t_k \leq T_k,k\in[K]
\Bigr\}$.
Training datasets are constructed by executing \(\Phi^g_k\) in the embodied environment
\(\mathcal{E}\) which returns a Bernoulli outcome $r^g_k \;=\; \mathcal{E}\bigl(g,\Phi^g_k\bigr)\in\{0,1\}$, indicating task failure or success.
Our objective is to learn a \(V_{\theta}\) that, with high probability, ranks promising solutions above unsuccessful ones. Concretely, we minimize the expected negative log-likelihood of selecting a successful solution:
\begin{equation}
    \min_{\theta}
\;\mathbb{E}_{g\sim{G}}
\Bigl[-\log \Pr\bigl(r^g_{\star}=1 \mid V_{\theta}(g,\Phi^g_{\star})\bigr)\Bigr],\;\text{where}\;\Phi^g_{\star} = \mathop{\arg\max}\limits_{\Phi^g_k \in \Omega(g)} V_{\theta}(g, \Phi^g_k).\label{eq:formulation}
\end{equation}
Two key challenges emerge: 1) execution of a reasoning trace (i.e., generating a single data point) is computationally and time inefficient for dataset construction; 2) failed solutions may contain useful scene configuration or subtask supervision.
To address these issues, we propose training the verifier at the granularity of consecutive subtasks, labeling any subtask that appears in a successful solution as positive. This enables the verifier to efficiently assess intermediate steps, according to their potential to contribute to task success, regardless of the outcome of the task.

\vspace{-0.12in}
\section{Methodology}
\vspace{-0.1in}

\begin{figure*}[htbp]
    \vspace{-0.15in}
    \includegraphics[width=\textwidth]{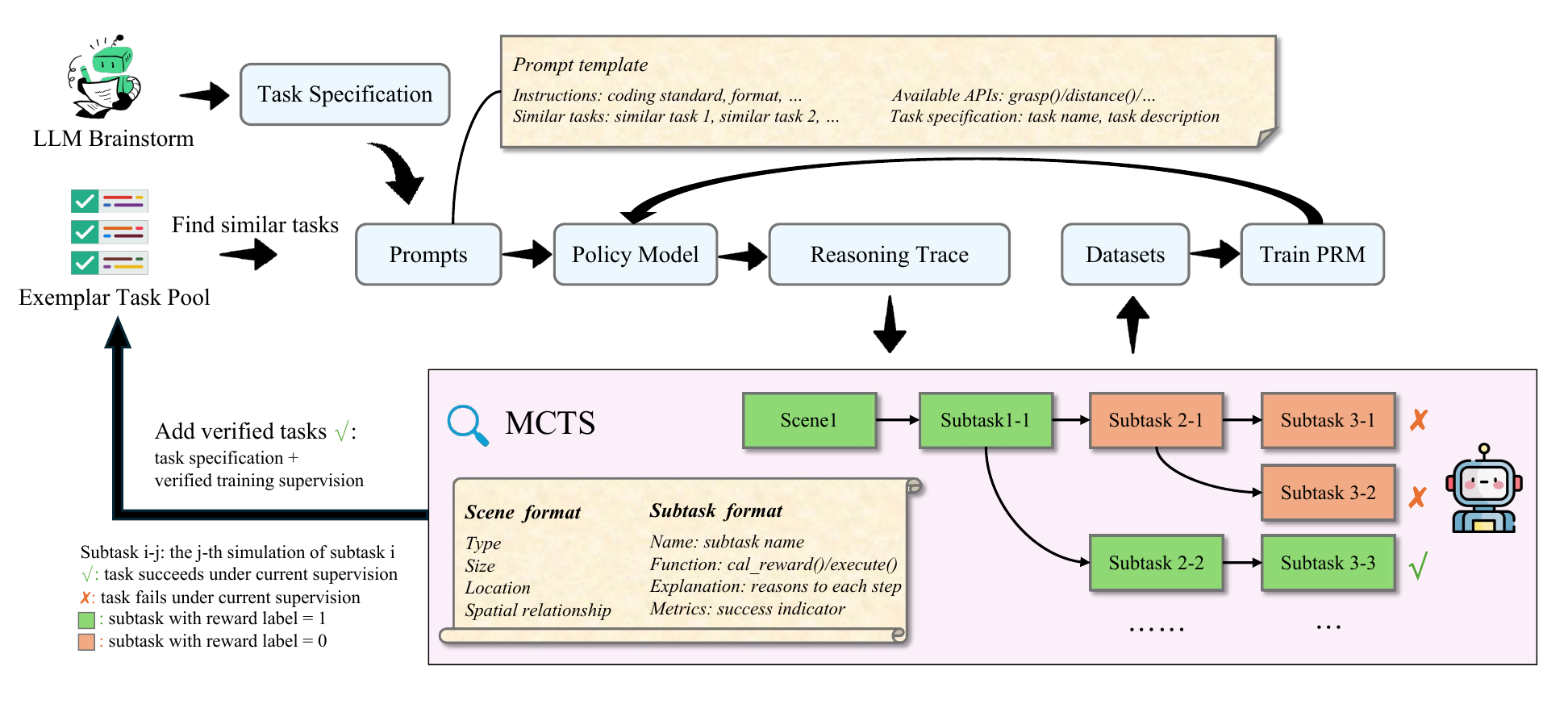}\vspace{-0.12in}
    \caption{The VERGSA framework.}
    \label{fig:framework}
    \vspace{-0.1in}
\end{figure*}

We propose VERGSA as a novel framework that leverages the verification model to rigorously validate the critical reasoning parts of generative skill acquisition. 
These include scene configuration, task decomposition into subtasks, generation of subtask training supervision, and success indicator formulation. 
The overall framework is illustrated in Figure \ref{fig:framework} and is detailed in the following.

\vspace{-0.13in}
\subsection{Exemplar Task Pool}
\begin{figure}[htbp]
  \centering
  \vspace{-0.1in}
  \begin{minipage}[t]{0.48\textwidth}
    \vspace{7pt}                          
    \centering
    \renewcommand\arraystretch{1.5}
    \captionof{table}{Top 2 similar tasks per novel task.}\vspace{-0.05in}
    \label{tab:similar-tasks}
    \resizebox{\linewidth}{!}{%
      \begin{tabular}{c|c c}
        \toprule
        \textbf{Novel Tasks} & \textbf{Similar Tasks}       & \textbf{Similarity Scores}\\
        \midrule
        Change Fan Direction & Change Lamp Direction (1st)  & 0.897\\
                             & Rotate Fan Rotor (2nd)       & 0.677\\
        \rowcolor{gray!20} Open Pot Lid & Remove Pot Lid  (1st) & 0.829\\
        \rowcolor{gray!20}             & Open Toilet Lid (2nd)      & 0.772\\
        Adjust Display Angle & Tilt Display Screen (1st)   & 0.909\\
                             & Rotate Laptop Screen (2nd)  & 0.824\\
        \bottomrule
      \end{tabular}%
    }
  \end{minipage}\vspace{-0.1in}
  \hfill
  \vspace{-0.1in}
  \begin{minipage}[t]{0.48\textwidth}
    \vspace{0pt}                          
    \centering
    \includegraphics[width=\linewidth]{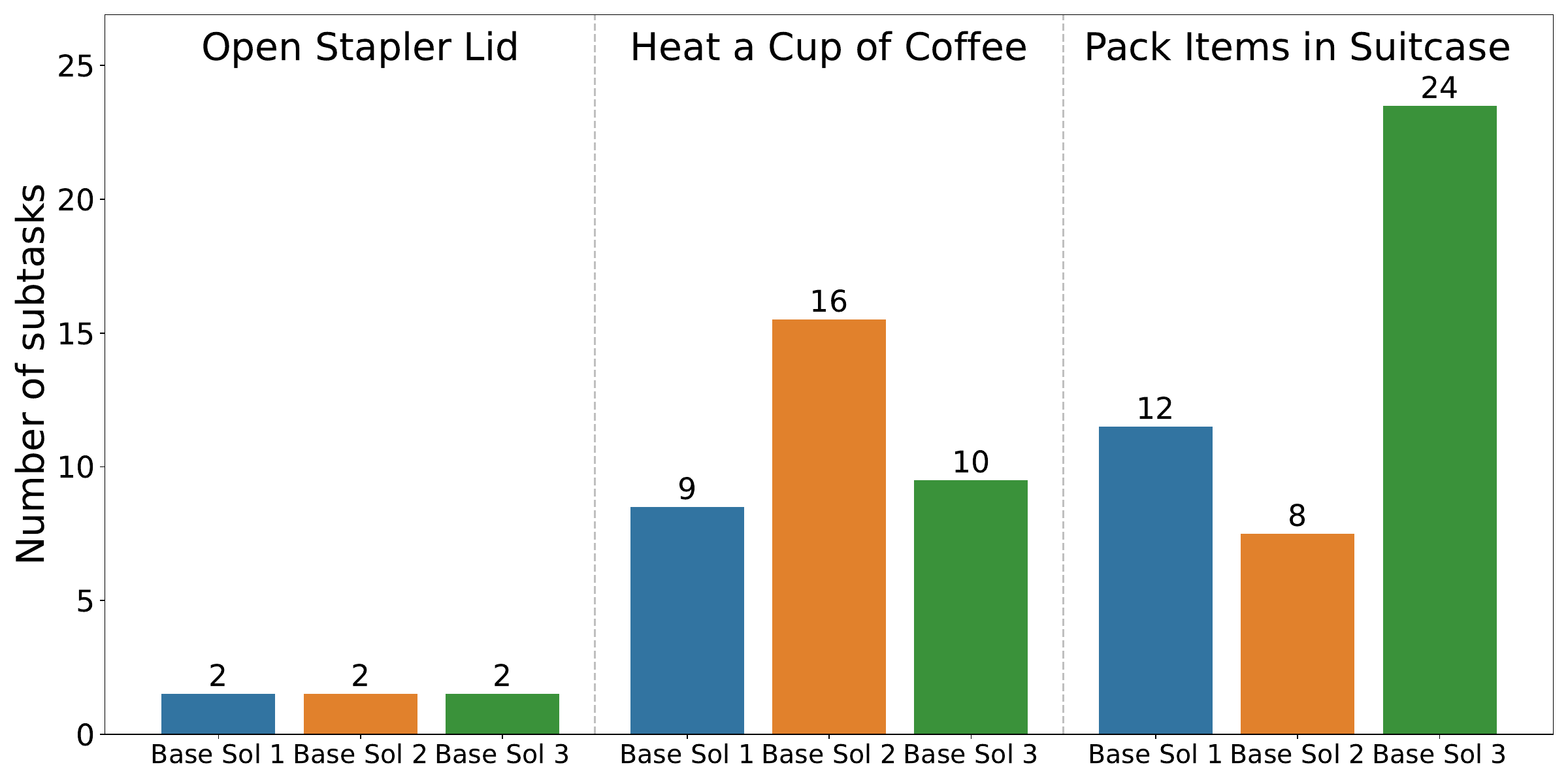}\vspace{-0.05in}
    \captionof{figure}{Number of subtasks in three tasks.}
    \label{fig:num-of-subtasks}
  \end{minipage}\vspace{-0.1in}
\end{figure}

\vspace{-0.08in}
Rather than using a fixed exemplar‐task template, we maintain a dynamic pool $\mathcal{T}$ of successfully executed tasks for LLM (policy model) prompting, whose subtask decompositions and subtask training supervisions provide valuable insights for tackling contextually similar, previously unseen tasks. For each task $ \tau_i \in \mathcal{T}$, the task name $n_i$ and description $d_i$ are extracted and then encoded into a single embedding vector using a pre-trained encoder $\Phi: \mathcal{X} \rightarrow \mathbb{R}^d$, such as Sentence-BERT. 
\begin{align*}
    \mathbf{e}_i = \Phi(n_i \oplus d_i),\; \forall \tau_i \in \mathcal{T}; \;\mathbf{e}_{\text{new}} = \Phi(n_{\text{new}} \oplus d_{\text{new}}),\; \text{where $\oplus$ denotes string concatenation. }
    \end{align*}
To retrieve relevant exemplars, we compute the cosine similarity between the new task embedding $\mathbf{e}_{\mathrm{new}}$ and each embedding in the pool, selecting the top-$K$ most similar tasks.
\begin{equation*}
    \omega_i = \frac{\mathbf{e}_{\text{new}} \cdot \mathbf{e}_i}{\|\mathbf{e}_{\text{new}}\| \|\mathbf{e}_i\|}, \quad 
    \mathcal{T}_{\text{result}} = \underset{\tau_i \in \mathcal{T}}{\mathrm{argTopK}}(\omega_i, K=2)\subseteq\mathcal{T}.
\end{equation*}
Upon successful execution of a novel task, it is appended to the pool, enabling continuous refinement of retrieval. This dynamic updating strategy maintains a robust and evolving resource for constructing prompt templates used in generation of subtask supervisions for novel tasks. Table \ref{tab:similar-tasks} reports the top two similar tasks from the pool for three novel tasks, highlighting similarities in either the action performed or the object involved. Table \ref{tab: complete-info-tasks} in the Appendix provides additional examples.
\vspace{-0.1in}

\subsection{Task Specification and Prompts}
\vspace{-0.07in}
In addition to a task name and a description, a novel task specification includes 1) an object articulation tree detailing links, joints, and semantics of links, sourced from the PartNetMobility dataset; 2) the initial task configuration, along with the relevant links and joints, provided by the task proposal step in RoboGen \cite{wang2024robogen}. The task specification, augmented with reference knowledge derived from the top similar tasks, available APIs provided by the simulated environment, and detailed instructions for generating subtask training supervisions, collectively form the input prompts that are fed into the policy model. This comprehensive input structure ensures the policy model is equipped with the necessary contextual and operational information to generate the output response accordingly. The policy model then breaks down the task into several subtasks and generates subtask-level training supervisions to guide learning in later task execution. Additionally, it produces success indicators that enable early termination when a subtask is accomplished.

\vspace{-0.12in}
\subsection{Training Supervisions} 
\vspace{-0.07in}
Long-horizon skill acquisition under a single holistic supervision remains challenging in the field \cite{mishra2023generative}. A common walkaround decomposes tasks into shorter-horizon subtasks, where generating supervision for subtask policy learning is tractable. Figure \ref{fig:num-of-subtasks} depicts the subtask counts across three solutions to three novel tasks. Appendix Table \ref{tab:detail-subtasks} and Appendix Figure \ref{fig:all-tasks} provide additional examples.
Subtask policy learning typically relies on one of two supervision paradigms:
reinforcement learning \cite{schulman2017proximal, haarnoja2018soft}, guided by a reward function, and
motion planning \cite{karaman2011sampling}, driven by an execution function that reasonably invokes pre-defined primitive APIs in the simulator.
Although gradient-based trajectory optimization \cite{xian2023fluidlab, 10.5555/3524938.3525665} is another option, it requires the support of differentiable simulation, which is currently limited; we therefore omit it from our study. Prior work (RoboGen \cite{wang2024robogen}) showed that reinforcement learning excels in contact-rich, interactive tasks, while motion planning is superior for collision-free navigation. Nevertheless, the real-time verification of a better reward function or execution function remains unexplored.

\vspace{-0.12in}
\subsection{Automated Reward Labeling via MCTS} 
\vspace{-0.07in}
In mathematical problem-solving, substep rewards are typically estimated via Monte Carlo Tree Search (MCTS) simulations. A policy model first generates $N$ base solutions $\{\Phi_{i,0,0}\}_{i=1}^N$ with decoded answers $\{\alpha_{i,0,0}\}_{i=1}^N$, where the second zero indicates the substep number and the third zero indicates the base solution. 
For the $j$-th ($j\geq 1$) substep of the $i$-th base solution $\phi_{i,j,0} \in \Phi_{i,0,0}$, a completer is applied to finalize subsequent substeps, resulting in $K-1$ completed solutions $\{\Phi_{i,j,k}\}_{k=1}^{K-1}$ with decoded answers $\{\alpha_{i,j,k}\}_{k=1}^{K-1}$. 
The reward for this substep is then calculated based on these $K$ answers. 
We observe that, for non-branching substeps, decoded answers carry informative reward signals, presenting an opportunity to improve sample efficiency in MCTS. In embodied tasks, success metrics are generated for each subtask and the overall task, eliminating the concern of false positive reasoning traces that arise in mathematical problem-solving, where intermediate substeps are incorrect but the overall solution succeeds. As a result, we additionally assign reward labels to subtasks generated by the completer. This approach is detailed in Algorithm \ref{alg:arlet-mcts}. 
\begin{table*}[!ht]
    \centering
    \vspace{-0.05in}
    \caption{Completed solutions via MCTS of base solution $2$ for task Knock On Door.}
    \label{tab:mcts}
    \resizebox{0.9\textwidth}{!}{
    \begin{tabular}{l l l l l}
        \toprule
        $\Phi_{2,0,0}$ & $\Phi_{2,1,1}$ & $\Phi_{2,1,2}$\\
        \midrule
        $\phi_{2,0,0,0}:$ move to the door & $\phi_{2,1,1,0}:$ move to the door & $\phi_{2,1,2,0}:$ move to the door\\
        \rowcolor{gray!20} $\phi_{2,0,0,1}:$ knock on the door & $\phi_{2,1,1,1}:$ grasp the door & $\phi_{2,1,2,1}:$ adjust position for knocking\\
        \rowcolor{gray!20} $\phi_{2,0,0,2}:$ return to initial position & $\phi_{2,1,1,2}:$ move the door slightly for the knock & $\phi_{2,1,2,2}:$ knock on the door\\
        \rowcolor{gray!20} / & $\phi_{2,1,1,3}:$ move the door back to closed position & $\phi_{2,1,2,3}:$ release the door\\
        \rowcolor{gray!20} / & $\phi_{2,1,1,4}:$ release the door & /\\
        \midrule
        $\Phi_{2,0,0}$ & $\Phi_{2,2,1}$ & $\Phi_{2,2,2}$\\
        \midrule
        $\phi_{2,0,0,0}:$ move to the door & $\phi_{2,2,1,0}:$ move to the door & $\phi_{2,2,2,0}:$ move to the door\\
        $\phi_{2,0,0,1}:$ knock on the door & $\phi_{2,2,1,1}:$ knock on the door & $\phi_{2,2,2,1}:$ knock on the door\\
        \rowcolor{gray!20} $\phi_{2,0,0,2}:$ return to initial position & $\phi_{2,2,1,2}:$ move to the door & $\phi_{2,2,2,2}:$ release the door\\
        \rowcolor{gray!20} / & $\phi_{2,2,1,3}:$ knock on the door & $\phi_{2,2,2,3}:$ move away from the door\\  
        \rowcolor{gray!20} / & $\phi_{2,2,1,4}:$ move back to initial position & /\\
        \bottomrule
    \end{tabular}}
\end{table*}
\vspace{-0.1in}

Table \ref{tab:mcts} presents a case study of base and completed solutions generated for the `Knock On Door' task, showing only the subtask names for brevity. By pairing each subtask with its training supervision and reward label, we assemble the datasets used to train the Process Reward Model (PRM)
Given the LLM's capability to relentlessly brainstorm novel tasks, this procedure yields a large and diverse dataset. Appendix Table \ref{tab: appendix-case-study} gives a detailed example of subtasks, their training supervisions, and reward annotations from the datasets we constructed. 
To our knowledge, this is the first training dataset specifically designed to facilitate generative skill acquisition. The dataset will be publicly available to support further research and development in the field.

\begin{algorithm}
\caption{Automated Reward Labeling for Embodied Tasks via MCTS (ARLET-MCTS)}\label{alg:arlet-mcts}
\begin{algorithmic}[1]
\Require An embodied task with specification
\Ensure Reward function $\mathcal{R}$
\State Initialize MCTS tree $\mathcal{G}$ with root $v_0$

\For{$i \gets 0 \text{ to } N-1$}

    \State Expand $\mathcal{G}$ with base solution $\Phi_{i,0,0}$
    \For{$\text{each subtask } \phi_{i,j,0} \in \Phi_{i,0,0}$}
        \For{$k \gets 1 \text{ to } K-1$}
            \State \!Complete solution $\Phi_{i,j,k}$ from subtask $\phi_{i,j,0}$\!
            \State \!Expand $\mathcal{G}$ with solution $\Phi_{i,j,k}$
            \State \!Execute the completed solution $\Phi_{i,j,k}$
        \EndFor
    \EndFor
\EndFor

\For{$\text{subtask } \phi_{i,j,k,m}$ from solution $ \Phi_{i,j,k}$}
    \If{$\textsc{Success}(\phi_{i,j,k,m}) \land \textsc{Success}(\Phi_{i,j,k})$}  
        \State $\mathcal{R}(\phi_{i,j,k,m}) \gets 1$ 
    \Else
        \State $\mathcal{R}(\phi_{i,j,k,m}) \gets 0$ 
    \EndIf
\EndFor

\State \Return $\mathcal{R}$

\end{algorithmic}
\end{algorithm}

\vspace{-0.12in}
\subsection{Process Reward Model (PRM)} 
\vspace{-0.07in}
The policy model generates task solutions by decomposing them into subtasks and providing training supervisions, yet it lacks a verification model to verify and select among competing solutions. The PRM acts as a `critic' to the `actor' policy model by assigning reward labels to subtasks and their supervision, evaluating their quality at each reasoning step.
A well-designed PRM elevates the likelihood of selecting solutions aligned with correct skill acquisition paths, thereby improving the policy model’s success rate in generative skill acquisition. Specifically, given a task specification $g \in G$ and a sequence of consecutive subtask supervisions $z_{1:m}\subseteq \Omega(g)$, PRM ($ G \times \Omega \rightarrow \{0,1\}$) assigns a zero or one reward to the sequence.
Here, $z_{1:m}$ can be derived from any solution $\Phi_{i,j,k}$ by truncating at the $m$-th subtask and reserving the $1:m$ inclusive sequence.
We train a PRM with the binary cross-entropy loss, given by:
\[
L = - \frac{1}{M} \sum_{m=1}^{M} \big[ y_m\hat{y} \log(p_m) + (1 - y_m\hat{y}) \log(1 - p_m) \big],\label{eq:bce}
\]
where $M$ is the number of subtasks in solution $\Phi_{i,j,k}$; \( p_m\) is the predicted probability of sequence $z_{i:m}$ assigned by the PRM; $y_m,\hat{y}\in\{0,1\}$ are the $m$-th subtask and overall task success indicators for the solution $z_m$ belongs to, so that $y_m\hat{y}$ together indicates the golden label of sequence $z_{1:m}$. 
Under proper training, the PRM is expected to effectively verify and select the successful solution from multiple decoded candidates.

\section{Experiments}
VERGSA is a simple and scalable framework that can train and utilize a verification model to optimize generative skill acquisition in embodied tasks.
Our experiments are structured to address three research questions (RQs): 
\begin{itemize}[leftmargin=1em]
    \item {\it RQ1 (Effectiveness of the exemplar task pool):} How effective is the exemplar task pool in providing scene configurations and subtask supervisions by incorporating top similar tasks?
    \item {\it RQ2 (Effectiveness of the verification model):} How effective is the verification model in embodied reasoning for generative skill acquisition tasks?
    \item {\it RQ3 (LLM-as-a-Judge as an alternative):} Can LLM-as-a-Judge substitute for a verification model in optimizing generative skill acquisition?
\end{itemize}

\subsection{Experimental Setup}
We use Qwen2.5-Coder-32B-Instruct as the policy model and the completer. We sample $3$ base solutions per task and finalize $3$ completed solutions for each subtask from one of the three base solutions. We have assembled $246$ task initiatives, either brainstormed by Qwen2.5-Coder-32B-Instruct or derived from the task names outlined in \cite{wang2024robogen}. After human curation, scene configurations and subtask supervisions are created for $118$ of these tasks, as detailed in Appendix Figure \ref{fig:all-tasks}.

The exemplar task pool initially contains $15$ tasks and continues to expand by adding successfully executed tasks as the VERGSA framework operates. The names and descriptions of initial tasks in the pool are given in Appendix Table \ref{tab:appendix-exemplar}.
VERGSA can generate an unlimited stream of tasks, each with multiple solutions composed of scene configurations and subtask supervisions. We constructed a dataset of $30$ tasks, comprising $150$ solutions, of which $83$ solutions succeeded, resulting in an Average Task Success Rate (ATSR) of $55.33\%$. Additionally, these solutions are decomposed into $287$ subtasks, $144$ of which succeeded, resulting in an Average Subtask Success Rate (ASSR) of $50.17\%$. Appendix Table \ref{tab:appendix-dataset} detailed task and subtask success rates per task in the dataset. We implement the PRM with Qwen2.5-1.5B, which is trained using a learning rate of $1 \times 10^{-5}$, a weight decay of $0.005$, and a cosine learning rate scheduler with $100$ warm-up steps. To mitigate class imbalance during training, we optimize with focal loss for Eq. (\ref{eq:bce}). Training and evaluation are carried out on two disjoint subsets of tasks.
Detailed settings for reinforcement learning and motion planning training are provided in Appendix Section \ref{sec:appendix-implementation}. 

\subsection{Evaluation Metrics and Baselines}
\noindent\textbf{Success Rate (SR).} 
We report two metrics: ATSR and ASSR. The ATSR measures the proportion of successful solutions:
$\text{ATSR} = \frac{|Successful\_solutions|}{|Total\_solutions|}$
while the ASSR measures the proportion of successful subtasks:
$\text{ASSR} = \frac{|Successful\_subtasks|}{|Total\_subtasks|}$.

\noindent\textbf{Average Number of Subtasks (Avg. Num. of Sub.).} It computes the average number of subtasks per evaluated task via dividing the total number of subtasks by the number of solutions evaluated.


\noindent\textbf{Exemplar Task Pool (E.T.P.).} {\it w/o E.T.P.} refers to baselines that generate scene configurations and subtask supervisions without seeking guidance from context-similar tasks in the exemplar task pool. Instead, these baselines rely on fixed exemplar tasks pre-embedded within the prompt template.

\noindent\textbf{Strategies represent the success probability of one task ($\Sigma,\,\prod,\,\min, \max,$ {\normalfont Last})}. When evaluating PRM on a task, it generates rewards for the scene configurations and each subtask supervision in every solution. Several strategies can be employed to group these subtask rewards into a final score representing a solution's predicted success probability. A solution is considered successful only if the final score meets or exceeds a predefined threshold. The strategies are outlined below.
\begin{itemize}[leftmargin=1.1em]

    \item $\Sigma,\prod,\min,\max$: The arithmetic mean, geometric mean, minimum and maximum of predicted success probability across all subtasks, $\frac{1}{n}\sum_{i=1}^{n} p_i$, $\sqrt[n]{\prod_{i=1}^{n} p_i}$, $\min(p_1, p_2, ..., p_n)$ and $\max(p_1, p_2, ..., p_n)$;




    \item $\text{Last}$: The predicted success probability of the final subtask, $p_n$.
\end{itemize}



\noindent\textbf{LLM-as-a-Judge (L.a.a.J).} The LLM-as-a-Judge module employs LLMs to directly assess the validity and correctness of scene configurations and subtask supervisions, serving a role analogous to the PRM. A tailored prompt (see Appendix Section \ref{appendix-sec: prompt-judge}) instructs the LLM to check if the code is syntactically and logically correct and adheres to a predefined set of allowed API functions. 
API functions, in this context, refer to the predefined and permissible functions or libraries available within the simulation environment, which are utilized when executing subtasks. 
The LLM judge receives a task specification along with a subtask supervision and outputs a ternary judgment ($0$, $0.5$, or $1$), corresponding to rejection, uncertainty, or acceptance, respectively. For baseline comparisons, we evaluate several leading LLMs, including GPT-4o, DeepSeek-V3, Claude-3.5-Sonnet, Gemini-2.0-Flash, and Qwen2.5-72B-Instruct. 


\subsection{Experimental Results}
This section presents the experimental results addressing the three research questions outlined above.

\subsubsection{Effectiveness of the exemplar task pool}
\begin{wraptable}{r}{0.5\textwidth}
  \centering
  \setlength{\intextsep}{0pt} 
  \vspace{-1.2em}
  \caption{Performance of BaseModel and GPT-4o with and without Exemplar Task Pool.}
  \label{tab:w/o-exemplar-task-pool}
  \vspace{-0.4em}
  \resizebox{\linewidth}{!}{
    \begin{tabular}{l | c c c}
      \toprule
      Method                         & ATSR ($\uparrow$) & ASSR ($\uparrow$) & Avg. Num. of Sub. \\
      \midrule
      BaseModel                      & 0.74              & 0.80              & 9.0               \\
      \rowcolor{gray!20}
      BaseModel–{\it w/o} E.T.P.           & 0.53              & 0.55              & 17.8              \\
      GPT-4o                         & 0.86              & 0.88              & 8.2               \\
      \rowcolor{gray!20}
      GPT-4o–{\it w/o} E.T.P.              & 0.81              & 0.84              & 15.2              \\
      \bottomrule
    \end{tabular}
  }\vspace{-1em}
\end{wraptable}
In analogy to how mathematical formulas support specific classes of problems, such as the quadratic formula for solving all quadratic equations, we demonstrate that the exemplar task pool plays a role in guiding unseen tasks to generate reasonable subtask supervisions. We select five tasks with distinct contexts for our experiments, with results reported in Table \ref{tab:w/o-exemplar-task-pool}. The inclusion of the exemplar task pool demonstrates considerable performance improvements across both models. Specifically, it increases ATSR by $21\%$ and ASSR by $25\%$ for the BaseModel Qwen2.5-Coder-32B-Instruct. The average number of subtasks to accomplish a task has decreased from $17.8$ to $9.0$.
Similarly, for another more advanced LLM GPT-4o, the exemplar task pool boosts ATSR by $5\%$ and ASSR by $4\%$. The average number of subtasks to accomplish a task has decreased from $15.2$ to $8.2$. These results highlight the effectiveness of the exemplar task pool in enhancing task and subtask success rates and reducing unreasonable subtasks for tasks across different models.


\subsubsection{Effectiveness of the verification model}
We evaluate the verifier under two partitions.
{\it Task-based splitting} divides the dataset by different tasks. This approach assesses the verification model’s ability to generalize to novel, unseen tasks. {\it Solution-based splitting} divides the dataset by solutions, where a portion of the solutions for each task is
assigned to the training set, and the remaining solutions are
reserved for the test set. This approach assesses the verification model’s performance on previously encountered tasks.

The results in Table~\ref{tab:prm-compare} show that the PRM consistently improves the BaseModel’s performance across datasets split by both approaches. BaseModel-PRM variants outperform the BaseModel in ATSR and ASSR. Under task-based splitting, BaseModel-PRM-last achieves an ATSR and ASSR of $0.91$, compared to the BaseModel’s $0.67$. Under solution-based splitting, BaseModel-PRM-last reaches an ATSR of $0.92$ and ASSR of $0.97$ against the BaseModel’s $0.56$ and $0.62$. These results demonstrate PRM’s effectiveness in guiding novel tasks and refining similar tasks.

Strategies for grouping subtask rewards impact performance differently. The $\Sigma$ and $\prod$ strategies offer balanced results, while the $\min$ strategy, targeting the weakest subtask, achieves higher success rates in task-based splitting. The $\max$ strategy yields the lowest scores among all PRM varaints, and the $last$ strategy excels in both  partitions --- task-based (ATSR$=0.91$, ASSR $=0.91$) solution-based (ATSR $=0.92$, ASSR $=0.97$) --- underscoring the pivotal role of the final subtask.

\begin{table*}[!ht]
    \centering
    \renewcommand\arraystretch{1.5} 
    \caption{Quantitative results of PRM-verified subtask supervisions across datasets.}
    \label{tab:prm-compare}
    \resizebox{\textwidth}{!}{
    \begin{tabular}{l | c c c | l | c c c}
        \toprule
        Model / Dataset split by {\it tasks} & ATSR ($\uparrow$) & ASSR ($\uparrow$) & Avg. Num. of Sub. ($\downarrow$) & Model / Dataset split by {\it solutions} & ATSR ($\uparrow$) & ASSR ($\uparrow$) & Avg. Num. of Sub. ($\downarrow$) \\
        \midrule
        BaseModel &0.67 &0.67 &2.14 & BaseModel &0.56&0.62&2.42 \\
        
        \rowcolor{gray!20} BaseModel-PRM-$\Sigma$ &0.85     &0.85     &2.08       & BaseModel-PRM-$\Sigma$ &0.91&0.96&$\bm{2.27}$\\
        BaseModel-PRM-$\prod$ &0.85     &0.85     &2.08       &  BaseModel-PRM-$\prod$ &0.91&0.96&$\bm{2.27}$\\
        \rowcolor{gray!20} BaseModel-PRM-$\min$ &0.89     &0.89     &2.11       & BaseModel-PRM-$\min$ & 0.91 & 0.96 & $\bm{2.27}$ \\
        BaseModel-PRM-$\max$ &0.80     &0.81     &$\bm{2.07}$       & BaseModel-PRM-$\max$ &0.81     &0.92     &2.37       \\
        \rowcolor{gray!20} BaseModel-PRM-last &$\bm{0.91}$     &$\bm{0.91}$     &2.09       & BaseModel-PRM-last &$\bm{0.92}$     & $\bm{0.97}$     &2.41       \\
        \bottomrule
    \end{tabular}}
\end{table*}
\begin{table*}[!ht]
    \centering
    \renewcommand\arraystretch{1.5} 
    \caption{Quantitative results of LLM-as-a-Judge as an alternative to the verification model.}
    \label{tab:prm-compare-gpt4o}
    \resizebox{\textwidth}{!}{
    \begin{tabular}{l | c c c | l | c c c}
        \toprule
        Model / Dataset split by {\it tasks} & ATSR ($\uparrow$) & ASSR ($\uparrow$) & Avg. Num. of Sub. ($\downarrow$) & Model / Dataset split by {\it solutions} & ATSR ($\uparrow$) & ASSR ($\uparrow$) & Avg. Num. of Sub. ($\downarrow$)\\
        \midrule
        GPT-4o & $\bm{0.78\pm0.04}$ & $0.80\pm0.03$  & $\bm{2.03\pm0.03}$ & GPT-4o & $\bm{0.65\pm0.06}$ & $\bm{0.70\pm0.06}$ & $\bm{2.21\pm0.21}$ \\
        \rowcolor{gray!20}Deepseek-V3 &$0.74\pm0.00$ &$0.76\pm0.01$ &$2.08\pm0.03$ & Deepseek-V3 &$0.60\pm0.04$ &$0.68\pm0.06$ &$2.43\pm0.02$ \\
        Claude-3-5-Sonnet & $\bm{0.78\pm0.00}$ & $\bm{0.81\pm0.00}$ & $2.06\pm0.00$ & 
        Claude-3-5-Sonnet &$0.64\pm0.02$ &$0.66\pm0.01$ &$2.46\pm0.03$ \\
        \rowcolor{gray!20} Gemini-2.0-Flash & $\bm{0.78\pm0.01}$ & $\bm{0.81\pm0.00}$ & $2.06\pm0.00$ &
        Gemini-2.0-Flash &$0.60\pm0.02$ &$0.66\pm0.01$ &$2.38\pm0.03$ \\
        Qwen2.5-72B-Instruct &$0.76\pm0.06$  & $0.78\pm0.07$  & $2.08\pm0.02$  & 
        Qwen2.5-72B-Instruct &$0.61\pm0.03$    &$0.66\pm0.04$   & $2.28\pm0.07$    \\
        \bottomrule
    \end{tabular}}
\end{table*}

\subsubsection{LLM-as-a-Judge as an alternative}
The emergent abilities of LLMs present a promising approach for their application as evaluators, a concept known as the LLM-as-a-Judge \cite{zheng2023judging}. In mathematical problem-solving, LLM-as-a-Judge is employed to evaluate whether a particular substep is reasonable. Subsequently, consensus filtering is applied to refine the label by rewarding substeps that potentially contribute to a correct answer and are deemed reasonable by LLM-as-a-Judge \cite{zhang2025lessons}. In skill learning tasks, however, success metrics are generated for each subtask, allowing us to check the outcomes of tasks directly without requesting LLM-as-a-Judge for this. We therefore benchmark LLM-as-a-Judge as a verifier baseline for candidate solutions against our verification model under the identical experimental settings. We find that although LLM-as-a-Judge is an effective verification approach for BaseModel, as evidenced by the results in Table~\ref{tab:prm-compare-gpt4o}.,
one of its perforance exceeds the PRM verifier.

\begin{figure*}[htbp]
    \includegraphics[width=\textwidth]{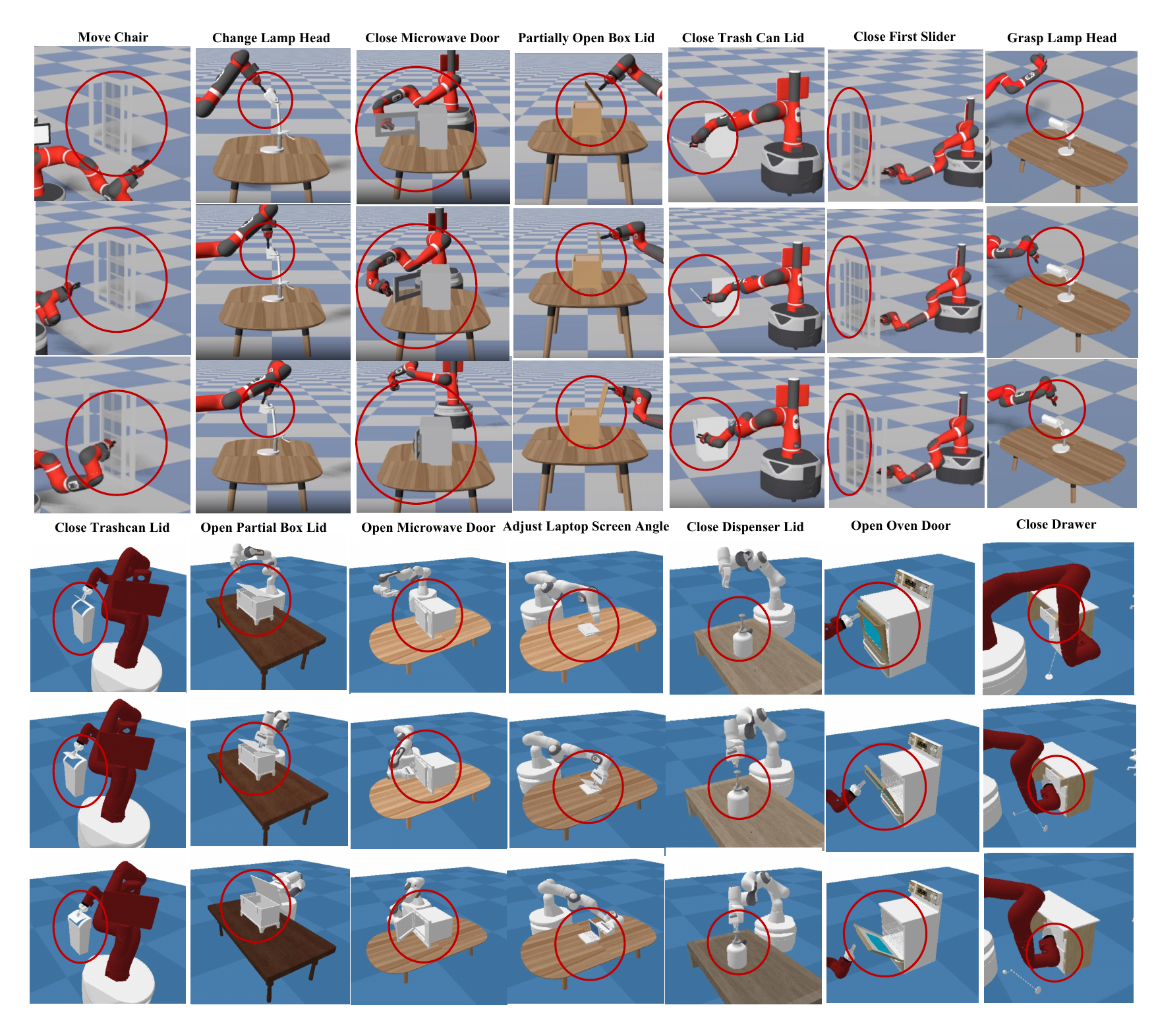}
    \caption{Performance of subtask execution across 14 tasks.}
    \label{fig:teaser2}
\end{figure*}

Figure \ref{fig:teaser2} presents qualitative results of intermediate snapshots from $14$ tasks, demonstrating the robotic arm’s skill acquisition via verified scene configuration and subtask supervision.

\section{Conclusion}

\noindent\textbf{Summary.}
We introduced {\it VERGSA}, a novel framework that formalizes the role of verification models in embodied reasoning for generative skill acquisition. VERGSA leverages an exemplar task pool to provide task-agnostic guidance for unseen scenarios while generating compositional success metrics for subtasks and overall task evaluations, providing feedback on generating per-subtask training supervisions. Additionally, VERGSA introduces a scalable reward labeling mechanism that automatically synthesizes dense reward signals for scene configurations and subtask supervisions. To the best of our knowledge, this approach yields the first comprehensive training dataset for verifying generative skill acquisition, establishing a solid foundation for real-time verification-driven skill learning.
Experiments validate that: 1) the exemplar task pool offers important guidance, improving ATSR by $21\%$; 2) the PRM verification model boosts success rates by $24\%$ for novel tasks and $36\%$ for encountered tasks; and 3) LLM-as-a-Judge with leading LLMs underperforms in skill acquisition tasks compared to the PRM verification model.

\noindent\textbf{Limitations and future directions.} 
The current framework has several limitations. First, complex skill acquisition is computationally inefficient, as it necessitates training a motion planning or reinforcement learning policy for each subtask based on a specific training supervision. This is further compounded by the computational overhead of running multiple simulations for base and completed solutions. 
Second, the translation of acquired skills from simulation to real-world applications is hindered by the sim2real gap due to discrepancies in physical dynamics, sensory noise, and actuator response between simulated and real environments. Mitigating this gap remains an open challenge in robotics.
Several future directions warrant further exploration. First, researchers could investigate meta-learning \cite{beck2023survey,gupta2018meta} and hierarchical reinforcement learning \cite{pateria2021hierarchical,nachum2018data} approaches to facilitate cross-task knowledge transfer or provide a warm-start base policy, thereby reducing per-subtask training overheads. Second, exploring domain randomization \cite{tobin2017domain,xu2023roboninja}, high-fidelity physics modeling \cite{li2020incremental,Genesis}, and hybrid training strategies \cite{10.5555/3524938.3525665,wang2023diffusebot} could enhance policy robustness for real-world deployment.

\clearpage
\bibliographystyle{IEEEtran}
\bibliography{ref}


\appendix
\onecolumn
\section*{Appendix}
\setcounter{section}{0} 
\renewcommand{\thesection}{\Alph{section}}
\section{Additional Implementation Details}\label{sec:appendix-implementation}
In this paper, we ran experiments on a server with a total of 8 NVIDIA GeForce RTX 4090 GPUs, each equipped with 24GB of memory.

We adopt the default settings of RoboGen \cite{wang2024robogen} for both the motion planning and reinforcement learning algorithms. For action primitives, RoboGen employs BIT${}^*$ \cite{gammell2015batch}, implemented in the Open Motion Planning Library (OMPL) \cite{sucan2012open}, as the motion planning algorithm. For grasping and approaching primitives, RoboGen first samples a surface point on the target object or link and then computes a gripper pose by aligning the gripper’s y-axis with the normal vector of the sampled point. The pre-contact gripper pose is positioned 0.03m above the surface point along the normal direction. Motion planning is then used to generate a collision-free path to the target gripper pose. Once the target pose is reached, the gripper continues moving along the normal until contact is established.
For reinforcement learning, RoboGen utilizes Soft Actor-Critic (SAC) \cite{haarnoja2018soft} as the algorithm. In object manipulation tasks, the observation space consists of the low-level state of the objects and the robot. Both the policy and Q networks in SAC are implemented as Multi-Layer Perceptrons (MLPs) with a size of $[256,256,256]$. The learning rate is set to $3\times10^{-4}$ for the actor, critic, and entropy regularizer. All manipulation tasks have a horizon of 100 steps, with a frameskip of 2. The action space of the RL policy is 6-dimensional: the first 3 elements define the translation (either as a delta translation or a target location, as suggested by the BaseModel or other LLMs), while the last 3 elements determine the delta rotation, expressed as a delta-axis angle in the gripper’s local frame. Each sub-task is trained for 1 million environment steps. 
Additionally, we incorporate an early stopping mechanism during the training of the reinforcement learning policy if the skill learning for a sub-task achieves success.

\newcounter{rownumc}
\newcommand{\rownumber}{\stepcounter{rownumc}\therownumc}

\begin{table}[htbp]
    \centering
    \setcounter{rownumc}{0}
    \caption {20 selected tasks with the average number of subtasks calculated and the top 2 similar tasks.}
    \label{tab: complete-info-tasks}
    \vspace{5pt}
    \renewcommand\arraystretch{1.5} 
    \resizebox{1\textwidth}{!}{
    \begin{tabular}{c l | c| c c }
    \toprule
         {\textbf{Row}} & {\textbf{Task Name}} &  {\textbf{Avg. Num. of Subtasks} $\uparrow$}  & \textbf{Similar Task 1 (1st)} & \textbf{Similar Task 2 (2nd)} \\
         \hline
        \rownumber & Close Trashcan Lid & 2 & Throw Trash Away & / \\
        \rowcolor{gray!20}
        \rownumber & Fold Chair & 2.33 & Unfold Chair & Tilt Chair Seat \\
        \rownumber & Open Kettle Lid & 2.33 & Lift Kettle by Handle & / \\
        \rowcolor{gray!20}
        \rownumber & Power on Printer & 2.67 & Turn On the Printer & Turn On Water Faucet \\
        \rownumber & Rotate Knife Blade & 4.33 & Open Scissors & Rotate Safe Knob \\
        \rowcolor{gray!20}
        \rownumber & Scroll Wheel  & 5 & Turn On the Printer & Open Toilet Lid \\
        \rownumber & Flush the Toilet & 5.67 & Open Toilet Lid & / \\
        \rowcolor{gray!20}
        \rownumber & Set Toaster Level & 7 & Close Dispenser Lid & Move Door Slightly Open \\
        \rownumber & Turn On Water Faucet & 7.67 & Adjust Water Flow Faucet & / \\
        \rowcolor{gray!20}
        \rownumber & Throw Trash Away & 9 & Close Trashcan Lid & / \\
        \rownumber & Deliver Package Through Door & 10.67 & Open Door & Close Door \\
        \rowcolor{gray!20}
        \rownumber & Store Object in Safe & 13 & Rotate Safe Knob & Hold Door Open \\
        \rownumber & Set Oven Timer & 14.67 & close the oven door & Turn On the Printer \\
        \rowcolor{gray!20}
        \rownumber & Turning On Coffee Machine & 15.33 & Pull Lever to Start Coffee Brewing & / \\
        \rownumber & Turn On the Printer & 17.7 & Press Start Button Dishwasher & / \\
        \rowcolor{gray!20}
        \rownumber & Heat a Cup of Coffee & 23 & Pull Lever to Start Coffee Brewing & Open Microwave Door \\
        \rownumber & Change Lamp Direction & 24 & Rotate Lamp Head & / \\
        \rowcolor{gray!20}
        \rownumber & Spin Chair & 25 & Tilt Chair Seat & Adjust Chair Position \\
        \rownumber & Start Washing Cycle & 27 & Open Washing Machine Door & Press Start Button Dishwasher \\
        \rowcolor{gray!20}
        \rownumber & Close Storage Furniture Door & 40.33 & Close Door & close the drawer of the table \\

        \bottomrule
        
    \end{tabular}
    }
\end{table}

\begin{table}[ht]
\centering
\caption{Monte Carlo rollouts of training supervisions for task Fold Chair.}
\label{tab: appendix-case-study}
\vspace{5pt}
\begin{tabular}{p{1.7cm}|p{12cm}}
\toprule
\textbf{Task} & Fold Chair \\
\midrule
\parbox{3cm}{\it rollout 1} 
\colorbox{negative}{\parbox{1.6cm}{subtask 1-1}} &  
\fontsize{8}{10}\selectfont\fontsize{8}{10}\selectfont\begin{verbatim}
subtask 1: grasp the seat
```primitive
rgbs, final_state = grasp_object_link(self, "FoldingChair", "link_1")  
success = check_grasped(self, "FoldingChair", "link_1")
```
\end{verbatim}
\\
\midrule
\parbox{3cm}{reward label} & 0
\\
\midrule
\parbox{3cm}{\it rollout 1} 
\colorbox{negative}{\parbox{1.6cm}{subtask 2-1}} &
\fontsize{8}{10}\selectfont\begin{verbatim}
subtask 2: fold the seat by lowering it to its folded position
```reward
def _compute_reward(self):
    # The reward encourages the end-effector to stay near the seat
    eef_pos = get_eef_pos(self)[0]
    seat_pos = get_link_state(self, "FoldingChair", "link_1")
    reward_near = -np.linalg.norm(eef_pos - seat_pos)

    # Access the joint state of joint_1 for folding the seat. 
    joint_angle = get_joint_state(self, "FoldingChair", "joint_1") 

    # The joint angle corresponding to the folded position
    folded_angle = get_joint_limit(self, "FoldingChair", "joint_1")[0]
    
    # The reward will be the negative difference between the current joint 
    # angle and the folded angle.
    diff = np.abs(joint_angle - folded_angle)
    reward_joint = -diff
    
    reward = reward_near + 5 * reward_joint
    success = diff < 0.15 * (get_joint_limit(self, "FoldingChair", "joint_1")[1] 
             - folded_angle)

    return reward, success
```
\end{verbatim}
\\
\midrule
\parbox{3cm}{reward label} & 0
\\
\midrule
\end{tabular}

\end{table}

\begin{table}[ht]
\centering
\begin{tabular}{p{1.7cm}|p{12cm}}
\midrule
\parbox{3cm}{\it rollout 2}
\colorbox{positive}{\parbox{1.6cm}{subtask 1-1}} &  
\fontsize{8}{10}\selectfont\begin{verbatim}
subtask 1: grasp the seat
```primitive
rgbs, final_state = grasp_object_link(self, "FoldingChair", "link_0")  
success = check_grasped(self, "FoldingChair", "link_0")
```
\end{verbatim}
\\
\midrule
\parbox{3cm}{reward label} & 1
\\
\midrule
\parbox{3cm}{\it rollout 2} 
\colorbox{positive}{\parbox{1.6cm}{subtask 2-1}} &  
\fontsize{8}{10}\selectfont\begin{verbatim}
subtask 2: lower the seat to fold the chair by actuating joint_0
```reward
def _compute_reward(self):
    eef_pos = get_eef_pos(self)[0]
    seat_pos = get_link_state(self, "FoldingChair", "link_0")
    reward_near = -np.linalg.norm(eef_pos - seat_pos)

    joint_angle = get_joint_state(self, "FoldingChair", "joint_0")
    joint_limit_low, joint_limit_high = get_joint_limit(self, "FoldingChair",
    "joint_0")

    # For folding, we want the seat to be lowered, so we target the lower 
    # joint limit
    diff = np.abs(joint_angle - joint_limit_low)
    reward_joint = -diff

    reward = reward_near + 5 * reward_joint

    # 10% tolerance of joint range
    success = diff < 0.1 * (joint_limit_high - joint_limit_low)  

    return reward, success
```
\end{verbatim}
\\
\midrule
\parbox{3cm}{reward label} & 1
\\
\midrule
\end{tabular}

\end{table}

\begin{table}[ht]
\centering
\begin{tabular}{p{1.7cm}|p{12cm}}
\midrule
\parbox{3cm}{\it rollout 3} 
\colorbox{positive}{\parbox{1.6cm}{subtask 1-2}} &  The same as {\it rollout 2} {subtask 1-1}
\\
\midrule
\parbox{3cm}{reward label} & The same as {\it rollout 2} {subtask 1-1}, which is 1
\\
\midrule
\parbox{3cm}{\it rollout 3} 
\colorbox{negative}{\parbox{1.6cm}{subtask 2-2}} &  
\fontsize{8}{10}\selectfont\begin{verbatim}
subtask 2: lower the seat to fold the chair
```reward
def _compute_reward(self):
    # The reward encourages the end-effector to stay near the seat
    eef_pos = get_eef_pos(self)[0]
    seat_pos = get_link_state(self, "FoldingChair", "link_0")
    reward_near = -np.linalg.norm(eef_pos - seat_pos)

    # Access the joint state of joint_0 for lowering the seat. 
    joint_angle = get_joint_state(self, "FoldingChair", "joint_0") 

    # The joint angle corresponding to the folded position of the seat
    folded_angle = np.pi  # Assuming the folded position is pi (180 degrees)

    # The reward will be the negative difference between the current 
    # joint angle and the folded angle.
    diff = np.abs(joint_angle - folded_angle)
    reward_joint = -diff
    
    reward = reward_near + 5 * reward_joint
    success = diff < 0.15 * (get_joint_limit(self, 'FoldingChair', 'joint_0')[1] 
              - get_joint_limit(self, 'FoldingChair', 'joint_0')[0])

    return reward, success
```
\end{verbatim}
\\
\midrule
\parbox{3cm}{reward label} & 0
\\
\bottomrule
\end{tabular}
\end{table}

\begin{table}[ht]
    \centering
    \caption{3 selected tasks with detailed subtask names of 3 base solutions.}
    \label{tab:detail-subtasks}
    \vspace{5pt}
    \renewcommand\arraystretch{1.5} 
    \resizebox{1.05\textwidth}{!}{
    \begin{tabular}{c|c c c}
        \toprule
        \textbf{New Tasks} & \textbf{Subtasks of Base Solution 1} & \textbf{Subtasks of Base Solution 2} & \textbf{Subtasks of Base Solution 3}\\
        \midrule
        Adjust Fan Speed & grasp the fan frame & locate and grasp the fan speed control knob & locate the fan s speed control knob\\
        (2,2,3 subtasks for base solutions 1,2,3)                     & rotate the control knob on the fan frame & rotate the fan speed control knob & grasp the speed control knob\\
                             & / & / & rotate the speed control knob to adjust fan speed\\
                             
        \rowcolor{gray!20} Insert Bread into Toaster & grasp the toaster hinge knob & move slider to open the toaster slots & move bread slice towards toaster opening\\
        \rowcolor{gray!20}  (9,8,6 subtasks for base solutions 1,2,3)            & rotate the hinge knob to open the toaster slots & slide the toaster slots open using the slider & grasp the bread slice\\
        \rowcolor{gray!20}             & move the bread slice above the toaster slot & pick up bread\_slice\_1 & move bread slice into toaster opening\\
        \rowcolor{gray!20}             & grasp the bread slice & insert bread\_slice\_1 into slot 1 & release bread slice into toaster\\
        \rowcolor{gray!20}             & insert the bread slice into the toaster slot & pick up bread\_slice\_2 & close the toaster door\\
        \rowcolor{gray!20}             & release the bread slice & insert bread\_slice\_2 into slot 2 & press the toaster knob\\
        \rowcolor{gray!20}             & move the end effector away from the toaster slots & move slider to close the toaster slots & /\\
        \rowcolor{gray!20}             & grasp the toaster hinge knob & slide the toaster slots closed using the slider & /\\
        \rowcolor{gray!20}             & rotate the hinge knob to close the toaster slots & / & /\\
        Close Storage Furniture Door & grasp the first drawer & grasp the first door link link\_0 & grasp the first door hinge\\
        (9,12,6 subtasks for base solutions 1,2,3)             & close the first drawer & close the first door link link\_0 & close the first door hinge\\
                     & release the first drawer & release the first door link link\_0 & release the first door hinge\\
                     & grasp the second drawer & grasp the second door link link\_1 & grasp the second door hinge\\
                     & close the second drawer & close the second door link link\_1 & close the second door hinge\\
                     & release the second drawer & release the second door link link\_1 & release the second door hinge\\
                     & grasp the door & grasp the third door link link\_2  & /\\
                     & close the door & close the third door link link\_2 & /\\
                     & release the door & release the third door link link\_2 & /\\                 
                     & / & grasp the fourth door link link\_3 & /\\
                     & / & close the fourth door link link\_3 & /\\
                     & / & release the fourth door link link\_3 & /\\
        \bottomrule
    \end{tabular}}
\end{table}

\clearpage

\begin{figure*}[htbp]
    \centering
    \includegraphics[width=0.9\textwidth]{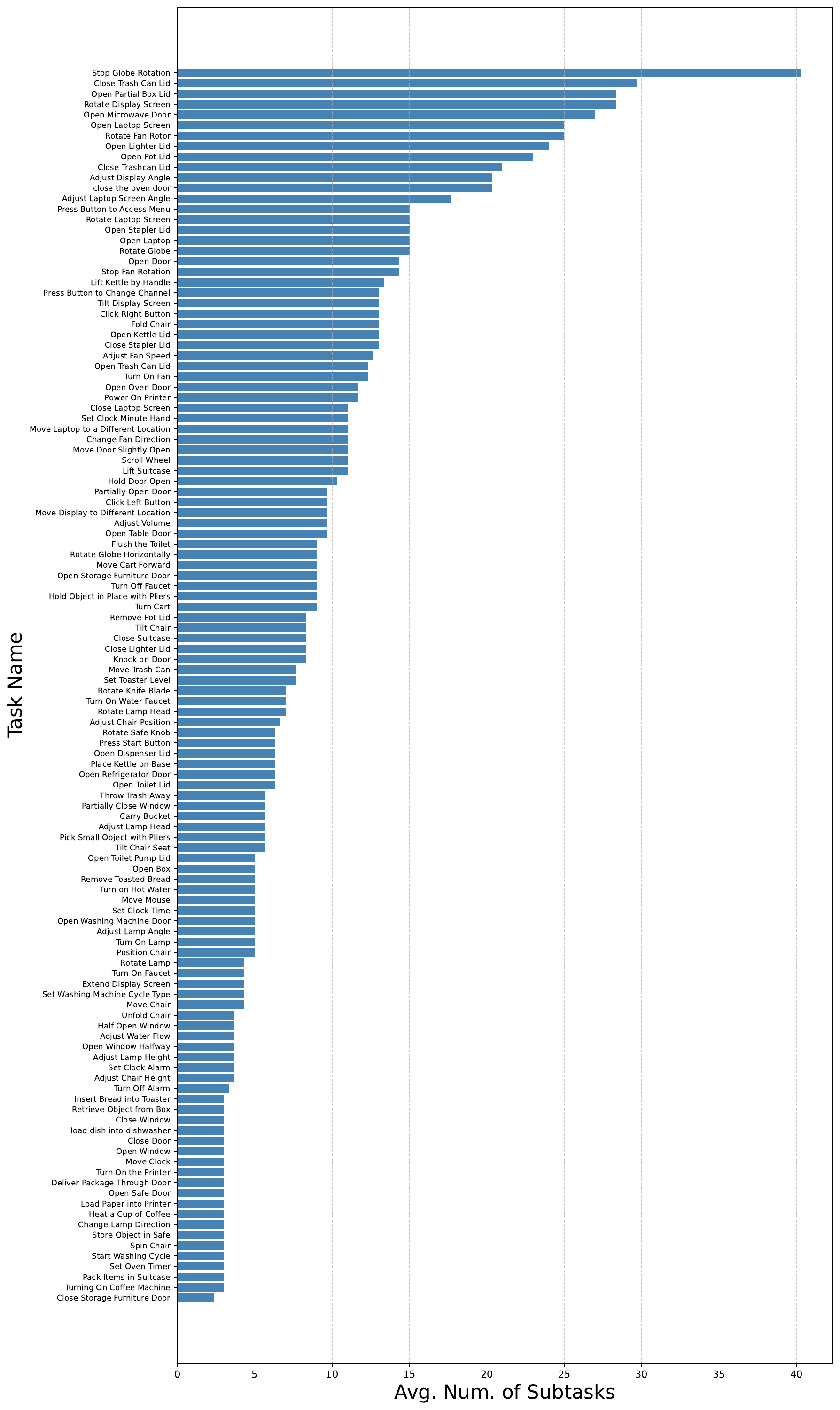}
    \caption{The distribution of the average number of substeps per solution for 118 selected tasks.}
    \label{fig:all-tasks}
\end{figure*}

\clearpage
\begin{table}[!ht]
\centering
\caption{Detailed task names and descriptions in the initial exemplar task pool.}
\label{tab:appendix-exemplar}
\vspace{5pt}
\setcounter{rownumc}{0}
\renewcommand\arraystretch{1.5} 
\resizebox{\textwidth}{!}{
\begin{tabular}{l|l}
\hline
\textbf{Task Name} & \textbf{Task Description} \\
\hline
Lift Kettle by Handle & The robotic arm approaches the kettle, grasps the handle and lifts the kettle off its base \\
\rowcolor{gray!20}Turn On Lamp & The task involves the robotic arm operating a lamp, specifically by turning it on This requires manipulating the switch to move it to the on position \\
Carry Bucket & The robot arm needs to pick up the bucket by its handle, carry it to the desired location, and then place it down safely \\
\rowcolor{gray!20}Flush the Toilet & The robot arm will actuate the button of the toilet in order to flush it \\
Close Dispenser Lid & The robot arm will close the lid of the dispenser This task involves moving the slider lids to a position indicating that the dispenser is closed \\
\rowcolor{gray!20}Turn Off Faucet & The robot arm needs to interact with the faucet to switch it off, stopping the flow of water \\
Adjust Water Flow & The robot arm needs to rotate the faucets switch in order to adjust the water flow Depending on the task, the robot may need to turn the water on, adjust the flow to a specific level, or turn the water off \\
\rowcolor{gray!20}Change Lamp Direction & The robotic arm adjusts the direction of the lamp by manipulating its head and rotation bars The goal is to change the illumination direction of the lamp \\
Open Window Halfway & The robot arm will manipulate the window to open it halfway and then stop, maintaining the position \\
\rowcolor{gray!20}Opening Both Refrigerator Doors & The robot arm will open both doors of the refrigerator \\
Turn on Hot Water & The robot arm manipulates the switch to turn on the hot water by actuating the toggle button \\
\rowcolor{gray!20}Pull Lever to Start Coffee Brewing & The robotic arm needs to pull a lever on the coffee machine to start brewing coffee \\
Adjust Chair Position & The task requires the robot arm to adjust the position of a chair, presumably by moving it to a new location or reorienting it \\
\rowcolor{gray!20}Turning On Coffee Machine & The robot arm has to turn on the coffee machine by manipulating the given lever and knobs \\
Press Start Button & The Franka Panda robotic arm will press the start button of the dishwasher \\
\hline
\end{tabular}
}
\end{table}

\begin{table}[!ht]
\centering
\caption{Detailed task and subtask success rates per task in the dataset}
\label{tab:appendix-dataset}
\vspace{5pt}
\renewcommand\arraystretch{1.5} 
\resizebox{\textwidth}{!}{ 
\begin{tabular}{l|c|c|c|c}
\toprule
\textbf{Task Name} & \textbf{Task Success Rate (\%)} & \textbf{Successful Tasks / Total} & \textbf{Subtask Success Rate (\%)} & \textbf{Successful Subtasks / Total} \\
\midrule
Adjust Display Angle & 100.00 & 6/6 & 100.00 & 9/9 \\
\rowcolor{gray!20}Adjust Fan Speed & 0.00 & 0/3 & 0.00 & 0/7 \\
Adjust Laptop Screen Angle & 83.33 & 5/6 & 88.89 & 8/9 \\
\rowcolor{gray!20}Close Stapler Lid & 50.00 & 2/4 & 37.50 & 3/8 \\
close the oven door & 0.00 & 0/3 & 0.00 & 0/6 \\
\rowcolor{gray!20}Close Trashcan Lid & 60.00 & 3/5 & 50.00 & 4/8 \\
Fold Chair & 25.00 & 1/4 & 33.33 & 3/9 \\
\rowcolor{gray!20}Move Door Slightly Open & 50.00 & 3/6 & 46.15 & 6/13 \\
Move Laptop to a Different Location & 25.00 & 1/4 & 33.33 & 4/12 \\
\rowcolor{gray!20}Open Door & 0.00 & 0/3 & 0.00 & 0/6 \\
Open Kettle Lid & 83.33 & 5/6 & 72.73 & 8/11 \\
\rowcolor{gray!20}Open Laptop Screen & 80.00 & 4/5 & 75.00 & 6/8 \\
Open Lighter Lid & 0.00 & 0/3 & 0.00 & 0/6 \\
\rowcolor{gray!20}Open Microwave Door & 77.78 & 7/9 & 84.62 & 11/13 \\
Open Partial Box Lid & 75.00 & 6/8 & 81.82 & 9/11 \\
\rowcolor{gray!20}Open Pot Lid & 83.33 & 5/6 & 88.89 & 8/9 \\
Open Stapler Lid & 0.00 & 0/3 & 0.00 & 0/6 \\
\rowcolor{gray!20}Open Trash Can Lid & 57.14 & 4/7 & 66.67 & 10/15 \\
Power On Printer & 40.00 & 2/5 & 53.85 & 7/13 \\
\rowcolor{gray!20}Press Button to Access Menu & 0.00 & 0/3 & 0.00 & 0/6 \\
Press Button to Change Channel & 33.33 & 1/3 & 28.57 & 2/7 \\
\rowcolor{gray!20}Rotate Display Screen & 85.71 & 6/7 & 80.00 & 8/10 \\
Rotate Fan Rotor & 71.43 & 5/7 & 70.00 & 7/10 \\
\rowcolor{gray!20}Rotate Laptop Screen & 85.71 & 6/7 & 80.00 & 8/10 \\
Rotate Safe Knob & 50.00 & 3/6 & 36.84 & 7/19 \\
\rowcolor{gray!20}Scroll Wheel & 25.00 & 1/4 & 27.27 & 3/11 \\
Set Clock Minute Hand & 25.00 & 1/4 & 27.27 & 3/11 \\
\rowcolor{gray!20}Stop Globe Rotation & 25.00 & 1/4 & 33.33 & 2/6 \\
Tilt Display Screen & 80.00 & 4/5 & 66.67 & 6/9 \\
\rowcolor{gray!20}Turn On Fan & 25.00 & 1/4 & 22.22 & 2/9 \\
\bottomrule
\end{tabular}
}
\end{table}

\clearpage

\section{Prompt for LLM-as-a-Judge}\label{appendix-sec: prompt-judge}
\scriptsize
\begin{verbatim}
"""
You are an expert in robot task planning and execution. Evaluate the following subtask within the given task context 
and determine if it contains any errors based on the validation issues listed below.
Task context: {task_description}
Subtasks: {step_text}
## Validation Issues to Check
1. for **Primitive Subtasks**:
   - Missing `rgbs, final_state` assignment
   - Missing `success` condition
2. for **Reward-Based Subtasks**:
   - Missing reward components
   - Missing success condition
   - Improperly weighted reward components
3. **General Issues**:
   - Incorrect object/link/joint naming
   - Improper API usage or use functions not in the allowed APIs.
   - Any grammatical or logical errors
4. You do not have to: 
    - check for the completness of the subtasks,since many subtasks are partial.
---
## Allowed APIs
For Primitive Subtasks:
- `grasp_object(self, object_name)`
- `grasp_object_link(self, object_name, link_name)`
- `release_grasp(self)`
- `check_grasped(self, object_name, link_name=None)`
For Reward-Based Subtasks:
- `get_eef_pos(self)`
- `get_position(self, object_name)`
- `get_link_state(self, object_name, link_name)`
- `get_joint_state(self, object_name, joint_name)`
- `get_joint_limit(self, object_name, joint_name)`
- `get_bounding_box(self, object_name)`
- `get_bounding_box_link(self, object_name, link_name)`
- `in_bbox(self, position, bbox_low, bbox_high)`
For Both:
- Standard Python libraries (numpy, etc.)
- Mathematical operations and control flow
---
## Output Instructions
Output a single number: 
- `1` if the subtask has no errors based on the validation issues.
- `0` if the subtask contains any errors from the validation issues.
- Use `0.5` only if you are genuinely uncertain (e.g., insufficient information to judge).
Avoid intermediate values (e.g., 0.85) unless absolutely necessary; prefer outputs close to 0 or 1.
Output only the number, no text or explanations.
"""
\end{verbatim}
\normalsize


\end{document}